\documentclass[11pt,a4paper]{article}
\usepackage[hyperref]{emnlp2020}
\usepackage{times}
\usepackage{latexsym} 
\usepackage{amssymb}
\usepackage{amsmath}
\usepackage{booktabs}
\usepackage{url} 
\usepackage{enumitem}
\usepackage{mathtools}
\usepackage{fontawesome} 
\usepackage{contour}
\usepackage{pifont}

\setitemize{noitemsep,topsep=0pt,parsep=0pt,partopsep=0pt}

\usepackage{microtype}
\usepackage{color, colortbl}

\definecolor{Gray}{gray}{0.85}
\definecolor{LightCyan}{rgb}{0.88,1,1}
\aclfinalcopy 
\def\aclpaperid{3304} 

\title{\textit{Birds have four legs?!} \\ NumerSense: Probing Numerical Commonsense Knowledge\\ of Pre-trained Language Models}

\author{
Bill Yuchen Lin \qquad Seyeon Lee \qquad Rahul Khanna \qquad Xiang Ren\\
\texttt{\{yuchen.lin,seyeonle,rahulkha,xiangren\}@usc.edu}\\
Department of Computer Science,  \\ University of Southern California\\
}

\renewcommand\footnotemark{}

\begin{document} 
\maketitle

\begin{abstract}
Recent works show that pre-trained language models (PTLMs), such as BERT, possess certain commonsense and factual knowledge.
They suggest that it is promising to use PTLMs as ``neural knowledge bases'' via predicting masked words. 
Surprisingly, we find that this may not work for \textit{numerical commonsense knowledge} (e.g., a bird usually has \textit{two} legs).
In this paper, we investigate whether and to what extent we can induce numerical commonsense knowledge from PTLMs as well as the robustness of this process.
To study this, we introduce a novel probing task with a diagnostic dataset, \textsc{NumerSense}\footnote{\url{https://inklab.usc.edu/NumerSense/}}, containing 13.6k masked-word-prediction probes (10.5k for fine-tuning and 3.1k for testing).  
Our analysis reveals that: 
(1) BERT and its stronger variant RoBERTa perform poorly on the diagnostic dataset prior to any fine-tuning; 
(2) fine-tuning with distant supervision brings some improvement;
(3) the best supervised model still performs poorly as compared to human performance (54.06\% vs 96.3\% in accuracy).


\end{abstract}

\section{Introduction}
\label{sec:intro}

Pre-trained language models (PTLMs), such as BERT~\cite{Devlin2019BERTPO}, have yielded state-of-the-art performance on many natural language processing tasks.
Given PTLMs' cited ability to create  general, yet useful text representations, an investigation of their ability to encode commonsense knowledge into representations is warranted––commonsense knowledge is often required to have a full understanding of language.

Recently there have been a few recent works that do investigate the inquiry of whether PTLMs possess \textit{commonsense knowledge}~\cite{Petroni2019LanguageMA, Feldman2019CommonsenseKM, bouraoui2019inducing}.
\begin{figure}
	\centering
	\includegraphics[width=1\linewidth]{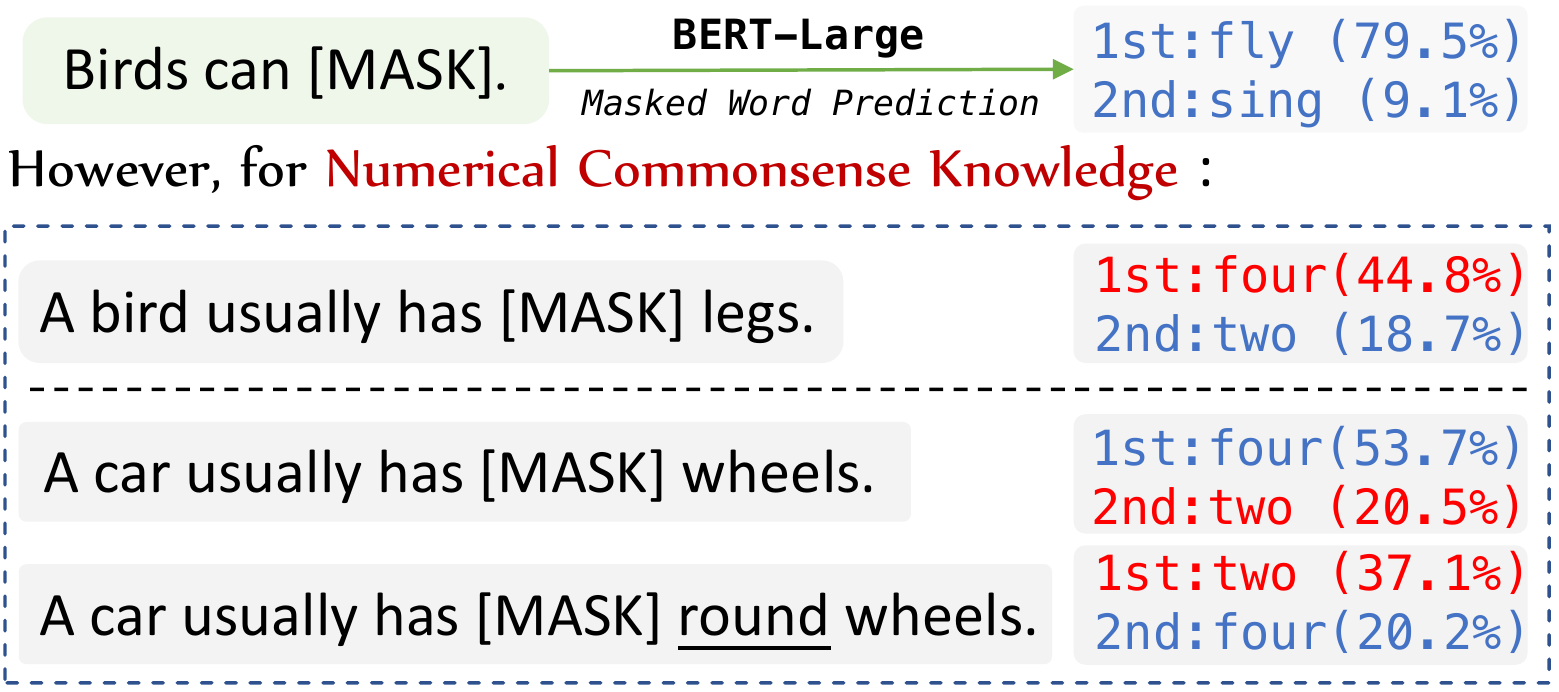}
	\caption{\small \textbf{Top:} PTLMs often cannot solve masked language modeling tasks needing \textbf{\textit{numerical commonsense knowledge}}, hence our title. \textbf{Bottom:} Even when PTLMs seemingly succeed, they fail to stay consistent under small perturbations.
	\vspace{-3em}
	}
	\label{fig:intro}
\end{figure}
Overall, these prior studies suggest that PTLMs are creating text representations that often have commonsense knowledge encoded in them. 
We, however, find it surprising that when posed with a similar reasoning-based masked-word-prediction task, PTLMs perform poorly in recalling the required \textit{numerical commonsense knowledge} (see Figure~\ref{fig:intro}). 

Therefore, in this paper,
our goal is to study whether PTLMs capture numerical commonsense knowledge, i.e., commonsense knowledge that provides an understanding of the numeric relation between entities.
We propose measuring this capability via a masked-word-prediction based probing task, where, the ranking of numeric words by what the model believes most probably fills the mask would expose the capabilities of PTLMs to capture \textit{numeric commonsense knowledge}. 
For example, the masked position in the sentence ``\textit{A bird usually has} \texttt{[MASK]} \textit{legs}.'' is best filled by the number ``two'' when considering only numerical words.

Around this concept, we built a carefully crafted dataset, \textsc{NumerSense}, of 3,145 probes that covers questions from 8 different categories such as everyday objects, biology, geometry, etc.
In our initial experiments, we find
PTLMs to be brittle against adversarial attacks. 
As shown in the bottom section of Figure~\ref{fig:intro}, BERT initially correctly predicts the masked word to be ``four'', but it changes its top result to ``two'' in the slightly perturbed second sentence (a simple insertion of the word `round'). 
Thus, we intentionally included adversarial examples in the probes to test the robustness.




We evaluate PTLMs in two settings (Section~\ref{sec:exp}): (1) a zero-shot setting, meaning no probes from our dataset were used to fine-tune the models before evaluation; (2) a distant supervision setting, where models were fine-tuned on examples from related commonsense reasoning datasets before being evaluated on ours.
Our findings reveal that PTLMs are still much worse than humans on the task, although fine-tuning with distant supervision can help.
We also provide some cursory analysis on why PTLMs perhaps preform so poorly, pointing to interesting future research. 
We also hope our work can benefit future works in: 1) improving PTLMs' abilities to faithfully capture (numerical) commonsense, 2) populating numerical facts in current commonsense knowledge bases, and 3) open-domain QA ––``Q: \textit{How many legs do ants have?'' ``A: Six!}''

	\section{The \textsc{NumerSense} Probing Task}
	\label{sec:numersense} 
	We introduce our numerical commonsense reasoning probing task, as well as the creation process of the namesake dataset, \textsc{NumerSense}.
	Then, we provide a breakdown of what types of knowledge are covered by the probes  and finally include additional high-quality distant supervision to test if fine-tuning   can improve performance.

	\subsection{Task Formulation}
	\label{ssec:task-f}
	We essentially probe PTLMs with the distribution of words a PTLM thinks could fill the masked position, 
	by ranking their softmax scores (greatest to least). 
	If the ranking demonstrates numerical commonsense knowledge––the highest ranked \textit{number word} (e.g., ``one'', ``two'', and so on) is the correct answer––then that probe is successfully completed by the PTLM. 
	The masked position in each probe is chosen such that a number word is an extremely probable way of filling in the blank.
		
	\subsection{Probing Data Collection}
		
	To build a suitable dataset for the proposed probing task, we make use of an existing corpus consisting of commonsense assertions, named \textit{Open Mind Common Sense} (OMCS)~\cite{singh2002open}.
	We first extracted the sentences from OMCS that had at least one of the following 12 \textit{number words}: \{``no''\footnote{ We include ``no'', as there exists statements involving numerical commonsense knowledge, where ``no'' is used in place of zero, ``There are \textbf{no} princes in the United States.''}, ``zero'', ``one'', ``two'', ..., ``ten'' \}.
	
	However, as to be expected, there were many noisy statements which were either 1) incorrect, 2) containing typos, or 3) having no numerical commonsense logic.
	We thus manually and pragmatically refined these sentences and did two rounds of vetting by different graduate students, from which we only kept the statements that were accepted by all annotators. 
	After this strict filtration process, we ended up \textbf{1,131} cleaned statements for probing.

	We did an initial test and observed that PTLMs can be brittle under a simple perturbation of inserting an adjective near the masked number word.
	Thus, in order to study the robustness of models in our proposed task, we also added adversarial examples to our dataset by adding adjectives before the noun involved in the numerical reasoning in each probe.
	The candidate adjectives are generated by querying relevant triples (e.g. $<$\texttt{wheel}, \texttt{HasProperty}, \texttt{round}$>$ for the example in Fig.~\ref{fig:intro}) in the commonsense knowledge graph, \texttt{ConceptNet}~\cite{Speer2017ConceptNet5A}, and further selected or modified by human annotators to assure adversarial examples are still valid and natural.
	We finally have \textbf{3,145} testing probes for \textsc{NumerSense} as the diagnostic dataset.
	
\begin{figure}
	\centering
	\includegraphics[width=1.\linewidth]{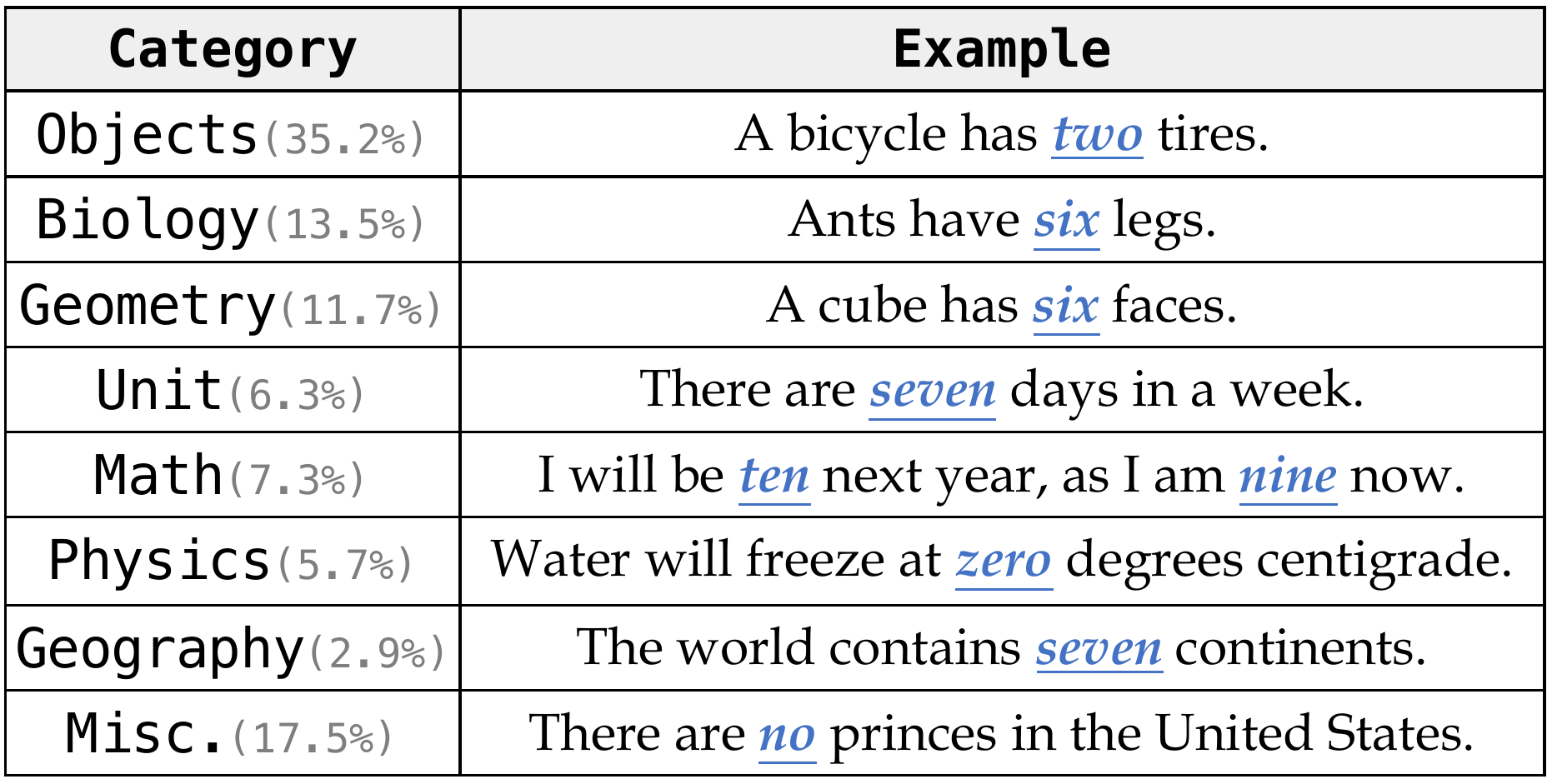}
	\captionof{table}{ \textsc{NumerSense} examples of each category.  
	}
	\label{fig:cate}
\end{figure}

	We also manually annotated the category label for each instance so that we can better understand the covered topics and their percentage. We found 8 types of numerical commonsense knowledge ranging from tangible everyday objects (e.g., car, guitar, and table) to geometry (e.g., cube). Table~\ref{fig:cate} lists some concrete examples of each category.

	\subsection{Supervision for Fine-Tuning PTLMs}
	\label{ssec:ds}
	One may wonder if fine-tuning towards this task could improve the performance. 
	In order to answer this question, 
	we further collected training sentences from the GenericsKB corpus~\cite{Bhakthavatsalam2020GenericsKBAK}.
	The sentences in GenericsKB are generic commonsense statements that are extracted from Simple Wikipedia, Common Crawl within educational domains, ARC corpus, etc.
	
	
	We collected these sentences by first obtaining a list of frequent nouns from various caption corpora such as MSCOCO~\cite{Lin2014MicrosoftCC} and VATEX~\cite{Wang_2019_ICCV}. 
	Then, we selected collected sentences contained at least one number word of interest and finally go through the same human annotator verification process as the test data. 
	We ended up collecting \textbf{10,492} sentences for fine-tuning and believe these sentences, if used properly, can improve PTLMs' ability to recall the numerical commonsense knowledge.

\subsection{Statistics of \textsc{NumerSense}}
\begin{figure}[h]
  \centering
  \includegraphics[width=1\linewidth]{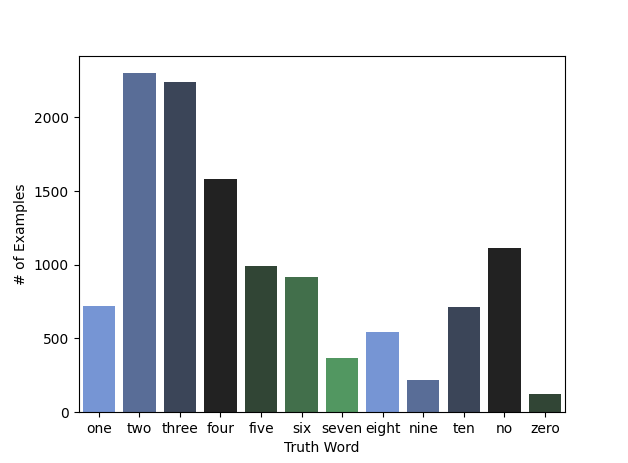}
  \caption{Truth number distribution of the training set.}
  \label{fig:train_dist}
\end{figure}

\begin{figure}[h]
  \centering
  \includegraphics[width=1\linewidth]{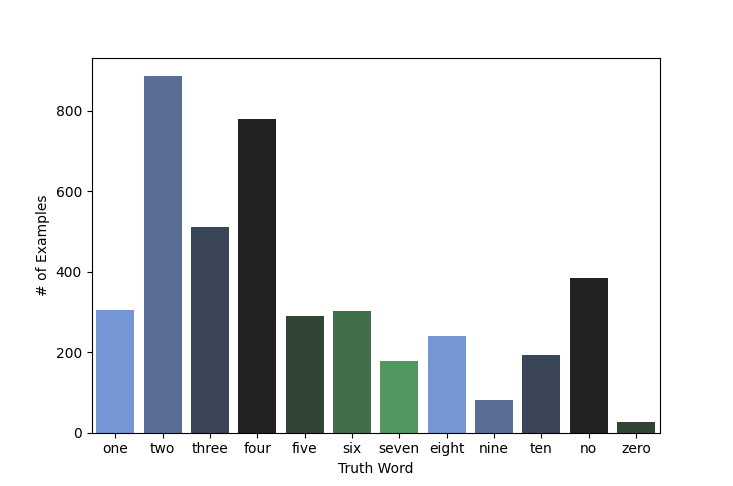}
  \caption{Truth number distribution of the test set.}
  \label{fig:test_dist}
\end{figure}

We show the distribution of the truth number words in the training data, test data in Fig.~\ref{fig:train_dist} and Fig.~\ref{fig:test_dist}.
The average length of the sentence in training data is 11.1 and it is 8.9 in test data.

\begin{table}[t]
\scalebox{0.68}{
\begin{tabular}{@{}c
>{\columncolor[HTML]{FFFFFF}}c 
>{\columncolor[HTML]{FFFFFF}}c 
>{\columncolor[HTML]{FFFFFF}}c |
>{\columncolor[HTML]{FFFFFF}}c 
>{\columncolor[HTML]{FFFFFF}}c 
>{\columncolor[HTML]{FFFFFF}}c @{}}
\toprule
\multicolumn{1}{l}{}                            & \multicolumn{3}{c}{\cellcolor[HTML]{FFFFFF}\textbf{\underline{Core Probes}} }                                     & \multicolumn{3}{c}{\cellcolor[HTML]{FFFFFF}\textbf{\underline{ + Adversarial Examples}}}                                \\ 
\multicolumn{1}{c}{\textbf{Models}}                            & \cellcolor[HTML]{FFFFFF}\textbf{hit@1} & \cellcolor[HTML]{FFFFFF}hit@2 & \cellcolor[HTML]{FFFFFF}hit@3 & \cellcolor[HTML]{FFFFFF}\textbf{hit@1} & \cellcolor[HTML]{FFFFFF}hit@2 & \cellcolor[HTML]{FFFFFF}hit@3 \\
\midrule
\cellcolor[HTML]{FFFFFF}GPT-2             & 29.86                         & 50.88                         & 67.49                        &   24.73   &   44.21  &     62.30              \\ \midrule
\cellcolor[HTML]{FFFFFF}BERT-Base               & 31.98                         & 55.92                         & 70.58                         & 25.24                         & 48.66                         & 64.81                         \\
\cellcolor[HTML]{FFFFFF}RoBERTa-Base            & 36.04                         & 60.42                         & 72.08                         & 28.39                         & 51.91                         & 67.29                         \\
\cellcolor[HTML]{FFFFFF}BERT-Large              & \underline{37.63}                         & 62.01                         & 76.77                         & 27.18                         & 52.89                         & 70.22                         \\
\cellcolor[HTML]{FFFFFF}RoBERTa-Large           & \textbf{45.85}                         & 66.70                         & 80.04                         & \textbf{35.66}                         & 58.52                         & 74.44                         \\
\midrule
\cellcolor[HTML]{FFFFFF}Ft. BERT-L.    & \underline{50.00}                         & 66.34                          & 74.91                        & 43.58                        & 62.27                        & 72.92                      \\
\cellcolor[HTML]{FFFFFF}Ft. RoBERTa-L. & \textbf{54.06}                         & 69.61                       & 79.15                        & \textbf{47.52}                         & 66.43                         & 76.76 \\
\midrule 
\cellcolor[HTML]{FFFFFF}\textit{Human Bound} & 
                    
\multicolumn{3}{c|}{\cellcolor[HTML]{EEEEEE}{{{89.7$^{(\alpha)}$ / 96.3$^{(\beta)}$ } }} }  
                   & \multicolumn{3}{c}{\cellcolor[HTML]{EEEEEE}{{{88.3 $^{(\alpha)}$ / 93.7 $^{(\beta)}$} }} }  
                        \\ \bottomrule
\end{tabular}
}
\caption{Results (\%) of PTLMs on \textsc{NumerSense}. `Ft.' stands for `Fine-tuned.' The human performance is shown by closed testing (${\alpha}$=`no external information') / open testing (${\beta}$=`Wikipedia is allowed'). \vspace{-1em}}
\label{tab:result}
\end{table}
\section{Empirical Analysis}
\label{sec:exp}
We introduce the set-up of the experiments and then present results from different PTLMs in both a zero-shot setting and a distantly supervised fine-tuned one.
We will also provide some analysis on the robustness and biases in the various models, and finally a study of the performance of a state-of-the-art open-domain question-answering model.
\subsection{Experiment Set-up}
We run our experiments in two settings, \textit{zero-shot inference} and additional supervision via fine-tuning. In the first setting, we probe PTLMs without any modifications, specifically we use BERT and RoBERTa with pre-trained masked-word-prediction heads.

In our second setting, we use our collected additional supervision dataset (Sec.~\ref{ssec:ds}) and mask the \textit{number words} in each sentence. We then proceed to fine tune the models above on these masked sentences, before evaluating them on \textsc{NumerSense}.



 \subsection{Evaluation Metric and Human Bound}
A masked-word-prediction head (either fine-tuned or not) produces a probability distribution over its whole vocabulary via a softax layer.
As mentioned in Sec.~\ref{ssec:task-f}, \textsc{NumerSense} is the task of using this probability distribution to rank all number words, and evaluating this ranking.
To evaluate,  we use hit@1/2/3 accuracy, which calculates the percentage of predictions where the correct number word is ranked in the top $k$ number words.\footnote{We also report the performance of GPT-2 by iteratively filling the masked word and rank with their perplexity. }

To estimate human performance on the task,
we sampled 300 examples and asked two groups of three people
to fill in the masked word, where one group had access to external information (\textbf{open-book} test) from the Web such as Wikipedia and the other did not (\textbf{closed-book} test).
We take the majority label as the final human label.

  	 \begin{figure}[t]
	\centering
	\includegraphics[width=1\linewidth]{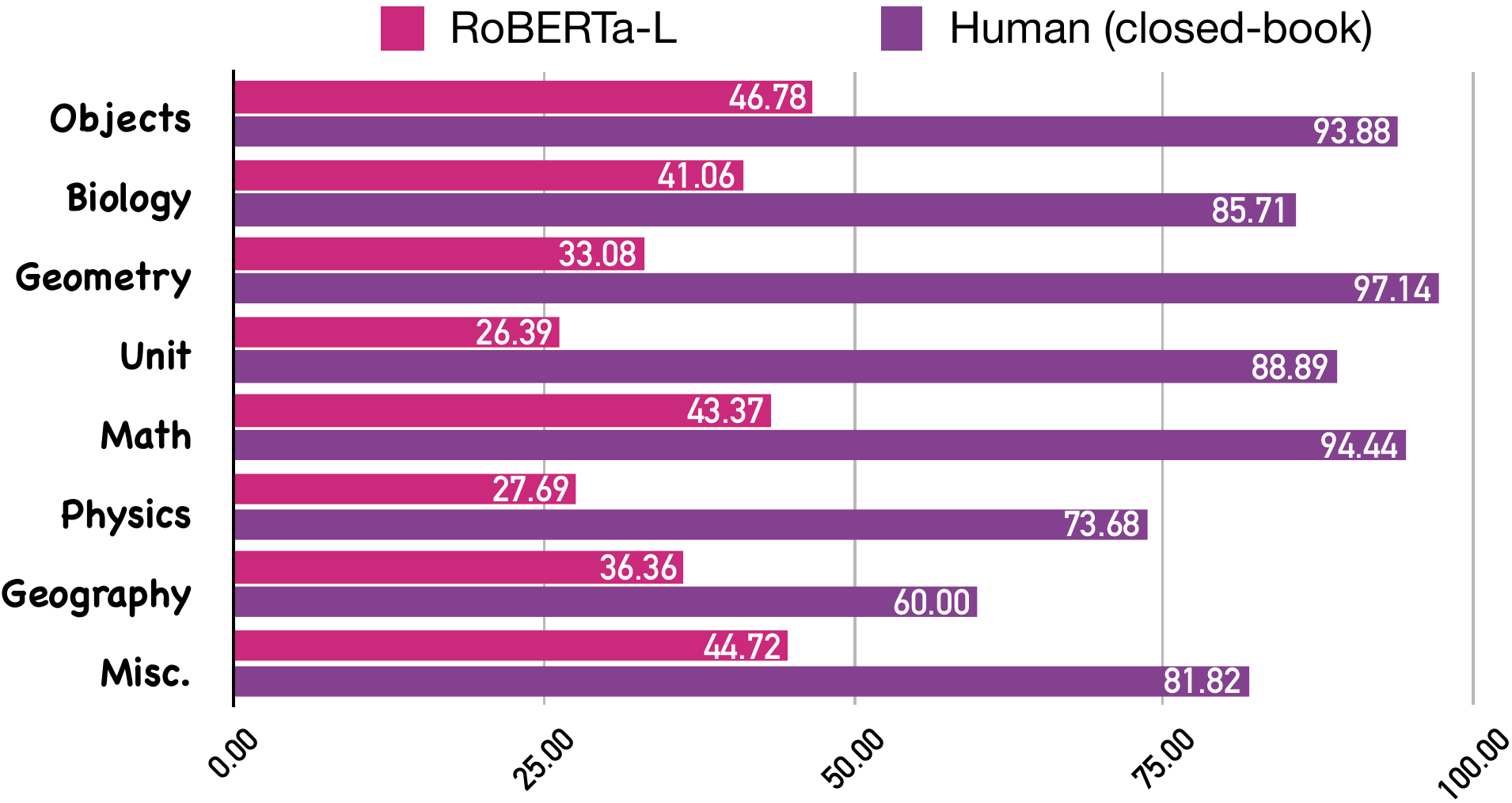}
	\caption{Performance of RoBERTa-Large V.S. human performance (closed-book tests) on  different categories of numerical commonsense knowledge. 
	}
	\label{fig:histcat}
\end{figure}
 
   \subsection{Experimental results}

\begin{figure*}[t]
	\centering
	\includegraphics[width=1\linewidth]{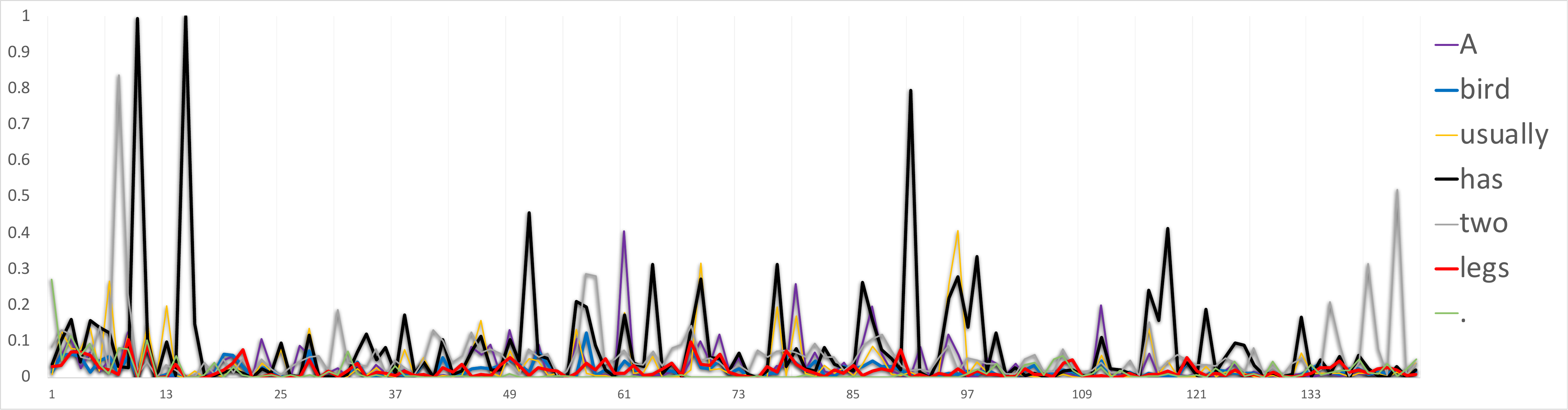}
	\caption{\small The attention distribution  of the sentence ``A bird usually has two legs.'' on RoBERTa-base.
	We plot the attention weights ($y$) between each word and the number word `two' at different position ($x$), e.g., $x=13$
	means (Layer 2, Head 1).
	\vspace{-1em}
	}
	\label{fig:attention}
\end{figure*}

  We show our experimental results in Table~\ref{tab:result}.
  The first four lines are results from PTLMs in the zero-shot inference setting. We see that size matters, as there is a clear performance gain when the model sizes increases.
  Also, RoBERTa's results are consistently better than BERT's, which is probably because RoBERTa uses a larger training corpora and focuses more on masked language modeling in its pre-training stage.
  
  We see that our fine-tuning efforts do help improve model performance: ``$37.63\rightarrow50.00$'' for BERT-large and ``$45.85\rightarrow54.06$'' for RoBERTa-large.
  However, both are still far from the human's closed-book evaluation.
  Figure~\ref{fig:histcat} shows PTLMs performance is poor across all categories within the core set of \textsc{NumerSense}.

  Comparing the performance of a PTLM on the ``Core Probes'' set (\#=1,131) versus the ``+ Adversarial Examples'' set (\#=3,145), we can measure their robustness.
  We found all models incur a significant performance drop when being evaluated on the adversarial set. 
  This suggests that PTLMs (even when fine-tuned) can be brittle towards adversarial attacks, and future direction in pre-training language models should consider more structured inductive biases such as dependencies and semantic roles when learning contextual representations.

\section{Case Studies}
\begin{figure}[t]
	\centering
	\includegraphics[width=1\linewidth]{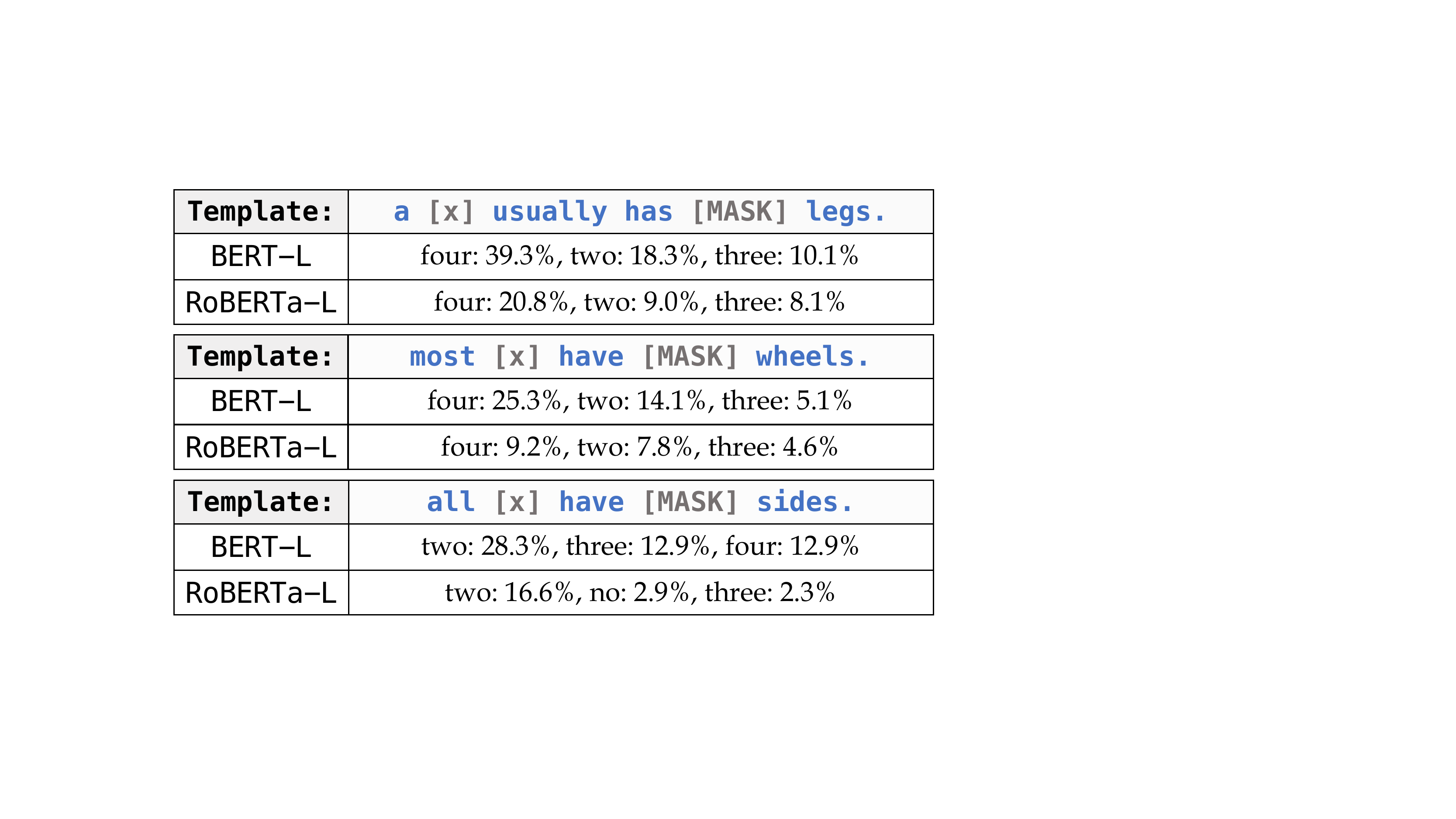}
	\captionof{table}{The average Softmax of top 3 predictions in templates where `[x]' is filled with 1k random words.
	\vspace{-2em}
	}
	\label{fig:bias}
\end{figure}

    \noindent
    \textbf{Object bias. }
	Recall the example ``a bird usually has \texttt{[MASK]} legs,'' which BERT-Large predicts to be ``four''.
	Does BERT-Large always predict ``four'' as long as the adjacent word after the [MASK] is `legs'?
	To investigate if the bias exists,
	we show some case studies in Table~\ref{fig:bias}. As 1,000 different randomly generated words fill the `[x]'s we see that both BERT and RoBERTa have a bias towards a certain answer, evidenced by the existence of a dominant answer in the softmax distribution. However, it seems that RoBERTa's ~\cite{Liu2019RoBERTaAR} modified pre-training strategy helps it have less bias. We argue that future studies should further control the bias in masked language modeling.

	\noindent
    \textbf{Attention distribution. }
    Following the prior probing work~\cite{Clark2019WhatDB} on the relationship between attention weights and syntactic structures, we plot the attention distribution of the sentence ``A bird usually has \textit{two} legs.'' with respect to the word `two' in Figure~\ref{fig:attention}.
    We find that the root word `has' enjoys the maximum attention at in the first few and middle layers, while the word `two' gets the maximum attention to itself in the end.
    The important words for querying the numerical commonsense, namely `birds' and `legs', always have low attention weights.
    This suggests that the BERT (and RoBERTa) may inherently lose the relationship between subject/object and number words.

\section{Open-Domain `How-Many' Questions}
	The examples in the \textsc{NumerSense} can be also seen as open-domain questions targeting `how-many' commonsense––``how many legs does a fly usually have?''
	Answering these open-domain numerical commonsense questions is a practical downstream application of models that are successful in the \textsc{NumerSense}.
	Thus, as a side note, we also report the performance of the state-of-the-art open-domain QA model~\cite{Asai2019LearningTR}.
	
	We use the model that is trained on the Natural Question (NQ) dataset~\cite{Kwiatkowski2019NaturalQA}, where we replace the `[MASK]'s in our examples with `how many', 
	so that our probes are in a similar format to NQ examples.
	For example ``a fly usually has [MASK] legs'' is converted to ``\textit{how many legs} a fly usually has?''\footnote{We also manually test some queries such as ``\textit{how many legs} does a fly usually have?'', which have similar results.} 
	The accuracy of the state-of-the-art model is only \textbf{15.4\%}, which is even lower than using BERT-base without fine-tuning.
	This indicates that improving performance on \textsc{NumerSense} can help improve the performance on answering open-domain ``how-many'' questions.

  

	\section{Related Work}
	\label{sec:relatedwork} 

\smallskip
\noindent 
\textbf{Probing Tasks for PTLMs.}
Prior work in probing language models have primarily focused on analysis of linguistic phenomena.
\citeauthor{Clark2019WhatDB} (2019) investigated the relationship between BERT's attention weights and syntactic structures, while
such as dependency (e.g. direct objects, noun modifiers), coreference, and sentence segmentation.
\citeauthor{Tenney2019BERTRT} (2019) was able to display where certain types of linguistic information is captured within BERT––they in fact find the layers in a PTLM  represent the steps of a classical NLP pipeline: POS tagging, parsing, NER, semantic roles, and coreference. 
This line of work has indeed helped us understand the ability of PTLMs to capture \textit{linguistic knowledge} via self-supervised learning from unlabeled data.
We are interested in the numerical commonsense knowledge of PTLMs.

\smallskip
\noindent 
\textbf{Probing Commonsense Knowledge.}
Besides the works that we have discussed in Section~\ref{sec:intro}, \citeauthor{Zhou2020EvaluatingCI} (\citeyear{Zhou2020EvaluatingCI}) and \citeauthor{Talmor2019oLMpicsO} (\citeyear{Talmor2019oLMpicsO}) also proposed to probe   the commonsense knowledge of pre-trained language models, following the prior work by~\citeauthor{Trinh2018ASM} (\citeyear{Trinh2018DoLM} and \citeyear{Trinh2018ASM}). 
They both utilized various existing language understanding datasets targeting commonsense knowledge to test if PTLMs can capture certain commonsense knowledge.
\citeauthor{lin-etal-2019-kagnet} (\citeyear{lin-etal-2019-kagnet}) also show that PTLMs can retrieve paths from ConceptNet that aid in interpreting the decision made by the PTLMs on the CommonsenseQA dataset~\cite{Talmor2018CommonsenseQAAQ}. 
\citeauthor{Lin2019CommonGenAC} (\citeyear{Lin2019CommonGenAC}) probe the commonsense knowledge in pre-trained language generation models via a constrained text generation task.
However, they do not consider numerical commonsense knowledge, which is relatively under-explored area.

\smallskip
\noindent
\textbf{Numerical Commonsense Knowledge.}
\citeauthor{forbes-choi-2017-verb} (\citeyear{forbes-choi-2017-verb}) and \citeauthor{goel-etal-2019-pre}~(\citeyear{goel-etal-2019-pre})
studied commonsense comparisons between two physical objects  (e.g., a house is usually bigger than a person) in pre-trained word embeddings.
\citeauthor{elazar-etal-2019-large}~(\citeyear{elazar-etal-2019-large}) and \citeauthor{yamane2020measuring} (\citeyear{yamane2020measuring})
propose to induce the commonsense distribution of  quantitative attributes (e.g., mass, length, and currency) of objects.
Their goal is to extract or crowd-source such numerical attributes, and then obtain distributions that reflect commonsense knowledge.
\textsc{NumerSense}, however, mainly focuses on exact numerical commonsense facts (e.g., a bird has \textit{two} legs) instead of a range of values (e.g., a tiger weighs \textit{around 120kg}), and have a larger number of arguments besides physical attributes. 

\smallskip
\noindent
\textbf{Encoding Numerics for Computation.}
\citeauthor{wallace-etal-2019-nlp} (\citeyear{wallace-etal-2019-nlp}) probe PTLMs in terms of the ability to represent numeracy tokens by a regression task (e.g., ``71'' $\rightarrow$ 71.0), and also find that BERT is not good at encoding numerical tokens.
Some works focus on incorporate algebra computation ability in PTLMs~\cite{zou-lu-2019-text2math, Geva2020InjectingNR}, thus making them able to answer  math reasoning tasks such as MAWPS~\cite{koncel-kedziorski-etal-2016-mawps} and DROP~\cite{dua-etal-2019-drop}.
Note that these models and tasks are not targeting numerical commonsense knowledge but mainly the numerical-related computation within text.



\section{Conclusion}
\label{sec:conclusion}  
We present a probing task, \textsc{NumerSense}, to induce numerical commonsense knowledge from pre-trained language models.
We collect a new diagnostic dataset carefully verified by human annotators, which covers 8 different topics.
Powerful pre-trained models such as BERT and RoBERTa perform surprisingly poorly, even after fine-tuning with high-quality distant supervision.
We hope our findings and probing dataset will provide a basis for improving pre-trained masked language models' \textit{numerical} and other concrete types of commonsense knowledge.

	\bibliographystyle{acl_natbib}
	\bibliography{acl2020}
	
	\clearpage

\end{document}